\documentclass[letterpaper, 10 pt, conference]{ieeeconf}  % Comment this line out if you need a4paper

\IEEEoverridecommandlockouts                              % This command is only needed if 
\usepackage{amsmath} % assumes amsmath package installed
\usepackage{graphicx}
\usepackage{booktabs}
\usepackage{mathtools}
\usepackage{bm}
\usepackage{tabularx}
\usepackage{caption,subcaption}
\usepackage{cite}
\usepackage{siunitx}
\usepackage{xcolor}
\usepackage{float} % for H in figure 
\usepackage{array,multirow}

\DeclareMathOperator{\sgn}{sgn}

\title{\LARGE \bf
Reinforcement Learning for Pivoting Task}

\author{Rika Antonova$^{1}$, Silvia Cruciani$^{1}$, Christian Smith and Danica Kragic% <-this % stops a space
\thanks{Rika Antonova, Silvia Cruciani, Christian Smith and Danica Kragic are with the Robotics, Perception and Learning Lab, CSC at KTH Royal Institute of Technology, Stockholm, Sweden.
{\tt\small\{antonova, cruciani, ccs, dani\}@kth.se}}%
\thanks {$^{1}$ Both of these authors contributed equally.}
}

\begin{document}
	
\maketitle
\thispagestyle{empty}
\pagestyle{empty}

%%%%%%%%%%%%%%%%%%%%%%%%%%%%%%%%%%%%%%%%%%%%%%%%%%%%%%%%%%%%%%%%%%%%%%%%%%%%%%%
%%%%%%%%%%%%%%%%%%%%%%%%%%%%%%%%%%%%%%%%%%%%%%%%%%%%%%%%%%%%%%%%%%%%%%%%%%%%%%%

\begin{abstract}
In this work we propose an approach to learn a robust policy for solving the pivoting task. Recently, several model-free continuous control algorithms were shown to learn successful policies without prior knowledge of the dynamics of the task. However, obtaining successful policies required thousands to millions of training episodes, limiting the applicability of these approaches to real hardware. We developed a training procedure that allows us to use a simple custom simulator to learn policies robust to the mismatch of simulation vs robot. In our experiments, we demonstrate that the policy learned in the simulator is able to pivot the object to the desired target angle on the real robot. We also show generalization to an object with different inertia, shape, mass and friction properties than those used during training. This result is a step towards making model-free reinforcement learning available for solving robotics tasks via pre-training in simulators that offer only an imprecise match to the real-world dynamics.
\end{abstract}

%%%%%%%%%%%%%%%%%%%%%%%%%%%%%%%%%%%%%%%%%%%%%%%%%%%%%%%%%%%%%%%%%%%%%%%%%%%%%%%
%%%%%%%%%%%%%%%%%%%%%%%%%%%%%%%%%%%%%%%%%%%%%%%%%%%%%%%%%%%%%%%%%%%%%%%%%%%%%%%
\section{Introduction}

In this work we address the problem of pivoting an object held by a robotic gripper. Pivoting consists of rotating an object or a tool between two fingers to reorient it to a desired angle. This problem falls in the general class of \emph{extrinsic dexterity} problems formulated by~\cite{dafle2014extrinsic}. Since many tasks in robotics require interaction with objects and tool use, the ability of placing items in the correct pose with respect to the gripper is important. 

Instead of relying on releasing the object and picking it up again \cite{tournassoud}, we focus on in-hand manipulation. Successful strategies for this dexterous task often rely on multi-fingered robotic hands or customized grippers \cite{trinkle, furukawa, chavan-dafle_1}. However, currently many robots have only parallel grippers, e.g. Baxter, PR2, Yumi. Thus it is important to develop regrasping strategies that do not rely on the additional degrees of freedom provided by complex hands.

We formalize the pivoting task as a reinforcement learning problem. The goal is to discover policies that, given the current state of the gripper and the tool, produce a sequence of actions to pivot the tool to the desired target angle. The actions consist of commanding the acceleration of the gripper and the distance for gripper's fingers. It is important to note that learning directly on the robot could focus too narrowly on the specific tool and robotic manipulator, while attempts to generalize would likely make the learning infeasible in terms of time. Hence it is not straightforward to directly apply reinforcement learning to this problem, if one aims to obtain policies that generalize beyond the exact hardware setup and tool properties.

\begin{figure}[t]
\centering
\includegraphics[height=0.28\textwidth]{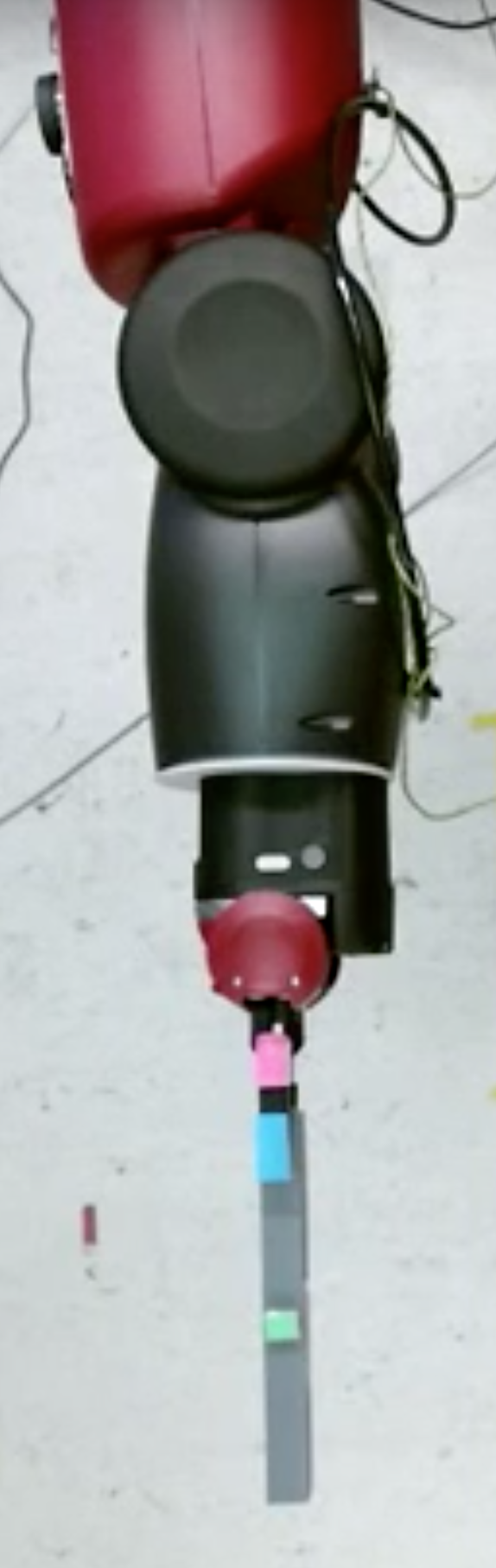}
\hspace{5px}
\includegraphics[height=0.28\textwidth]{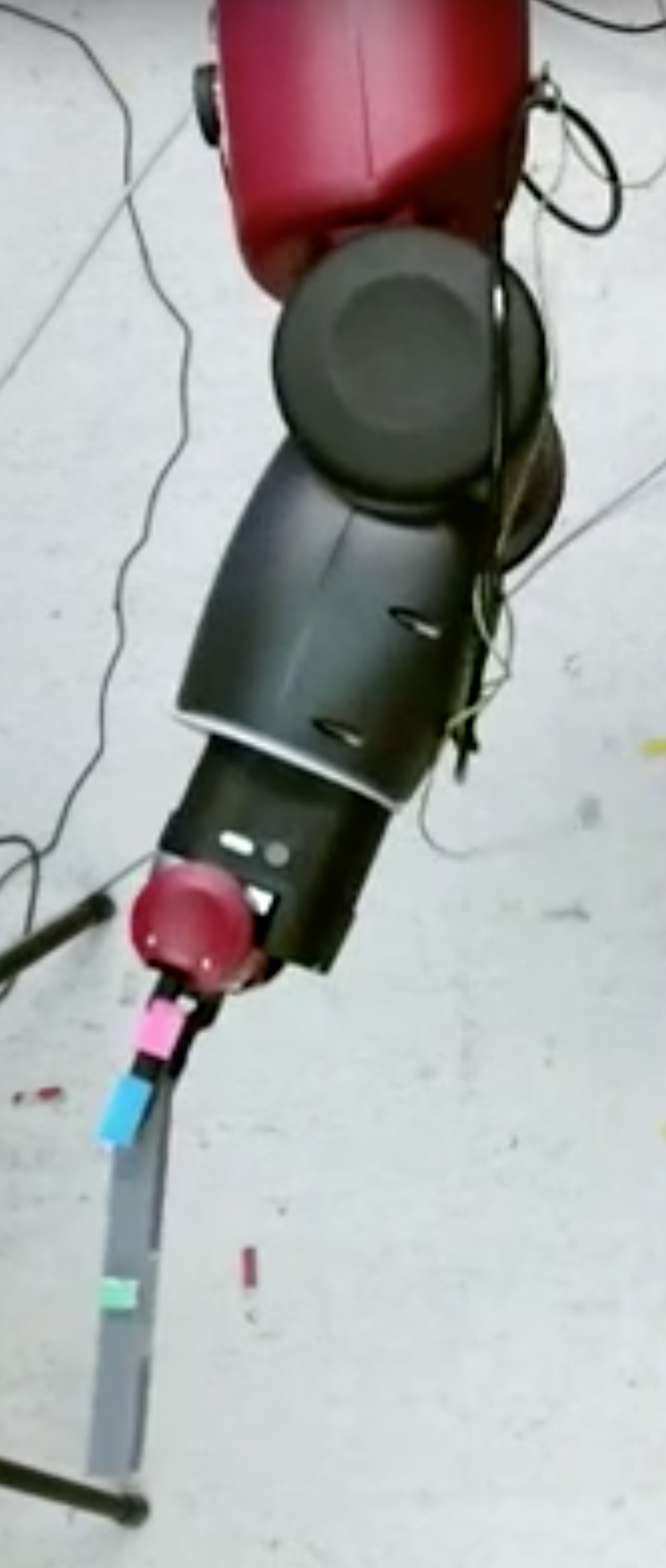}
\hspace{5px}
\includegraphics[height=0.28\textwidth]{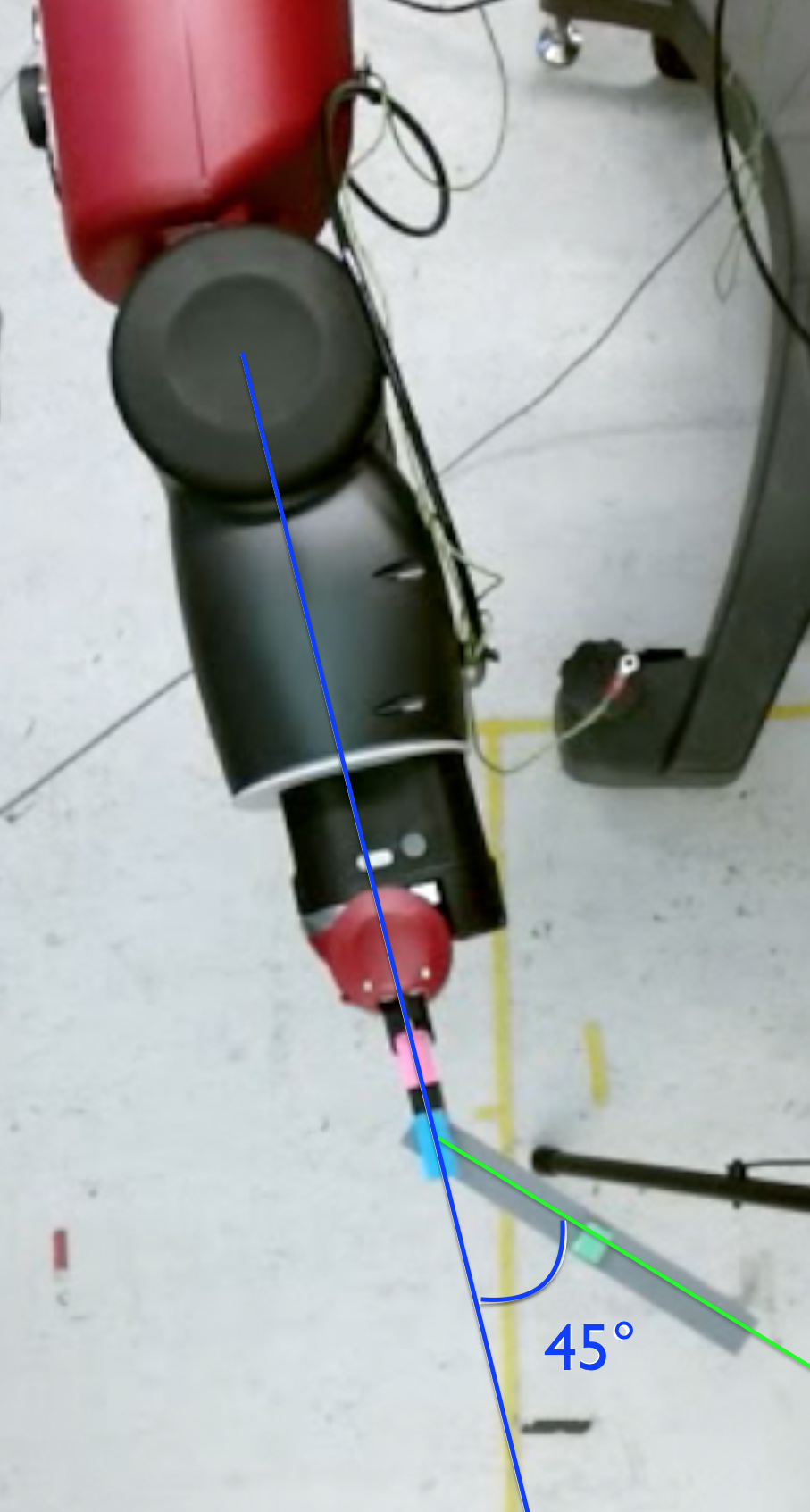}
\caption{\small Pivoting task executed on the Baxter robot -- pivoting the tool to $45^{\circ}$ with respect to the central axis of the gripper.}\label{fig_baxter}
\end{figure}

Our contribution is an approach to learn robust policies in a custom simulator, while taking into account the mismatch between the simulation and the real robot. We begin by constructing a simulator using dynamics equations for the setting. These equations describe well the general behavior of the system, but contain parameters that are infeasible to estimate precisely, like friction properties of the objects. Hence we generate a variety of training episodes with parameters randomly chosen from a range of values. We also simulate delays and errors in actuation. This results in learning policies robust to the mismatch between the outcome of control actions in simulation vs on a real robot. 
 
We designed this approach of training with a custom simulator so that we are not limited to only sample-efficient learning algorithms. While sample-efficient learning in real time can be effective, it could limit the flexibility and applicability of the learned policies to slightly altered hardware setups. In contrast, we can apply recent model-free policy search algorithms, even those not widely used for real robots previously. This facilitates learning flexible non-linear control policies, including those represented by deep neural networks. 

We demonstrate that our approach is able to learn policies to successfully solve the pivoting task when used on the robot. We present experiments with controlling the tool whose observable physical properties, like mass and inertia,  are used when constructing the simulator. We also demonstrate that the same policy is able to successfully control an object whose properties differ from those used during training. Fig.~\ref{fig_baxter} above shows one of our experiments with parallel gripper executing a pivoting task.

%==========================================================
\section{Related Work} \label{related_work}
\label{sec_relatedwork}

Extrinsic dexterity has been widely studied in robotics and is still an open challenge. Pivoting is one type of extrinsic dexterity problem that recently attracted attention of the robotics research community. In the following section we provide and overview of the previous work. Since in contrast to prior approaches ours employs deep reinforcement learning, we also provide a brief overview of this learning approach in the context of robotics.

%------------------------------------------------------------------------------
\subsection{Previous Work in Pivoting}
\label{subsec_relatedwork_dexterity}

Existing solutions for pivoting exploit environmental constraints, motions of the robot arm to generate inertial forces, and external forces, such as gravity. In \cite{holladay} the authors exploit gravity to rotate an object between two stable poses by using a contact surface. The pivoting is performed open loop and there is no control of the gripping force. Conversely, several other works on pivoting strongly focus on controlling the torque applied by the gripper on the object: in \cite{sintov} the authors focus on \emph{swing-up} motions. They address the problem using an energy-based control and they consider the ability of the gripper to exert dissipative torques on the object thanks to the friction at the pivoting point. In this case the motion of the object appears to be limited to a vertical plane and the approach strongly depends on fast sensory feedback and rapid response time of the gripper. 
The adaptive control for pivoting presented in \cite{vina_1} exploits gravity and controlled slip with only visual feedback. This approach has then been extended to consider also tactile feedback in \cite{vina_2}. However, the gripper is assumed to be in a fixed position, therefore the motion of the tool is determined only by the gravitational torque and the torsional friction. This motion is limited to be in a vertical plane and the proposed strategy can successfully reorient the object only when the desired configuration has less potential energy than the initial one. 

All these prior approaches rely strongly on having an accurate model of the object the gripper is holding, as well as precise measurement and modeling of the friction. Since it is difficult to obtain accurate estimates of these quantities, especially when they are related to friction modeling, in our work we do not rely strongly on highly accurate parameters for the successful outcome of a pivoting action.

%------------------------------------------------------------------------------
\subsection{Deep Reinforcement Learning in Robotics}
\label{subsec_relatedwork_rlrobotics}

Reinforcement Learning (RL) algorithms can be classified into two broad categories: model-based and model-free. Model-based algorithms can make learning data-efficient by assuming the dynamics of the task can be captured by a particular model (frequently parametric), then estimating the parameters of the model from samples obtained when learning to solve a task. However, strict global assumptions on the model class can limit the flexibility of the model-based representation. While this can be partially resolved by learning local models (e.g. ~\cite{atkeson1997locally}, ~\cite{levine2014learning}), such approaches might limit ability to incorporate prior knowledge -- for example, forgoing the knowledge that can be easily incorporated into dynamics equations in a simulator. 

Model-free RL algorithms instead allow to learn flexible control policies by directly interacting with either simulated or real environment. Recently, several model-free continuous state continuous action reinforcement learning algorithms have been proposed. Among these, two algorithms were reported to perform well on several simulated robotic tasks: Trust Region Policy Optimization (TRPO)~\cite{schulman2015trust} and Deep Deterministic Policy Gradient (DDPG)~\cite{lillicrap2015continuous}. Both use neural networks as policy and Q function approximators. While it has been shown that these approaches can handle problems similar in principle to those considered in robotics, significant practical problems arise when applying model-free policy search algorithms to real-world robotics tasks.

First of all, when using deep neural networks as function approximators, the question of data efficiency becomes key: the number of training episodes needed might be prohibitively large. For example, a recent benchmarking paper used up to 25 million steps when training~\cite{duan2016benchmarking}. The paper that introduced DDPG used up to 2 million steps, sometimes without achieving close to maximum reward (which in some cases means not being able to solve the task). Networks reported as successful had from 2 to 3 hidden layers with 25 to 400 nodes in each (15K to 200K parameters to learn, depending on the particular network structure). This was the case even for low-dimensional continuous problems, where state space is represented by joint angles, torques, velocities (as opposed to pixels) and the task of sensing the state is considered separate -- e.g. assumed to be performed by an off-the-shelf tracking system or achieved via custom hardware sensors.

Secondly, extensive hyper-parameter search might be needed for ensuring learning success on a given problem. In addition to the usual RL considerations, e.g. how aggressively to explore or how to shape the reward function, a new challenge is selecting the appropriate neural network architecture.

One additional challenge arises from the phenomenon known in deep learning literature as ``forgetting''. It has been observed that when training a neural network on a new task, the capacity to perform old tasks well can diminish. For the pivoting problem considered in this work, ``forgetting'' is problematic for several reasons. Reaching different target angles could be seen as slightly different tasks, since the optimal policies might differ. Theoretically, this is resolved by adding the target angle to the representation of the state. Practically, the knowledge acquired when training to reach different target angles is embedded in the same network, and this could impede or stall the learning progress during training. This presents a challenge for real-time learning, hence giving the motivation for our approach to learn a robust policy in simulation, where training on a large number of episodes could allow to recover from slow or inconsistent learning progress. The problem of ``forgetting'' in the context of learning in robotics is discussed at length in~\cite{rusu2016sim} -- the work that attempted to resolve the issue by pre-training in simulation, then using Progressive Neural Network architecture to continue learning on the real hardware. This or a similar approach could also be applicable to the pivoting task setting, however it might still require a significant amount of time for the second stage of training on the robot. Hence, we are motivated to develop an approach that can learn acceptable policies directly from the simulator, and would be applicable even when second stage training on the hardware is costly or infeasible.

%==========================================================
\section{Problem Description}
\label{sec_problem_description}

The problem we address is pivoting the tool to a desired angle while holding it in the gripper. This can be accomplished by moving the arm of the robot to generate inertial forces sufficient to move the tool and, at the same time, opening or closing the fingers of the gripper to change the friction at the pivoting point, gaining more precise control of the motion. 
% This can be accomplished by commanding changes in acceleration and gripper fingers to generate the forces causing the tool to rotate.
We assume that the robot is able to use one of the standard planning approaches to initially grasp the tool, but the position of the tool between the fingers is initially at some random angle $\phi_{init}$. The goal is to pivot the tool to a desired target angle $\phi_{tgt}$. 
%With such actions it is possible to control the tool to rotate between different angles. 

The first challenge is that the motion of the tool is influenced by the friction at the pivoting point. This is dictated by the materials of the tool and fingers, the deformation of the tool and fingers, and potentially the air flow. It is difficult to estimate precisely all the necessary coefficients to construct a high-fidelity friction model~\cite{vina_3}.

The second challenge is due to limitations of the robot hardware: delays in actuation, possible errors in execution, joint and velocity limits and constraints maintained to ensure safety. These cause the overall dynamics of the system to be uncertain. Combined with difficulties of estimating the object properties, this causes high uncertainty of the outcomes of the commanded actions overall.

The above challenges could be addressed by using a vision system with high frame rate to estimate the current state and high frequency control of the robot arm to ensure rapid response and re-adjustment. However, currently available commercial robots often have limited control frequency for adjusting gripper's fingers. Moreover, it is preferable to be able to solve the pivoting task with readily available vision systems, which have limited frame rate.

We take into consideration all of the challenges mentioned. Hence we propose a learning approach that lets us obtain control policies robust to some degree of uncertainty of both the tool properties and imprecisions of robot motion execution.

%==========================================================
\section{Modeling for Simulation}
\label{sec_system_model}
As mentioned in the previous section, simulators that can effectively incorporate global information about the task dynamics can be utilized for learning flexible control policies. In this section we describe the dynamic model of the system and the friction model that we used to simulate the pivoting task.

%------------------------------------------------------------------------------
\subsection{Dynamic Model}
\label{subsec_dynamic_model}
\begin{figure}[t]
\centering
\includegraphics[width=0.35\textwidth]{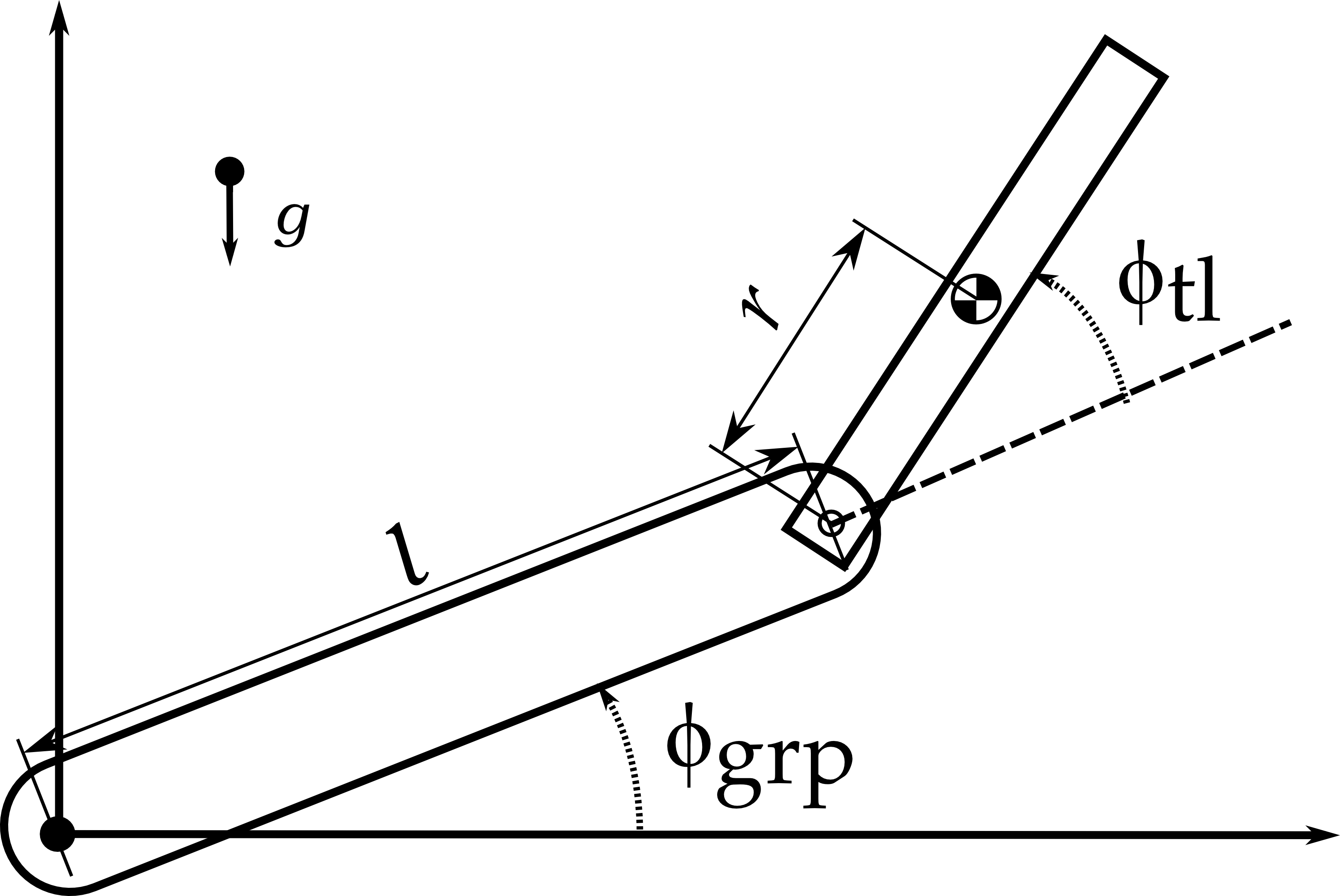}
\caption{\small Model of a 2-link planar arm. The first link represents the gripper and the second link represents the tool that rotates around the pivoting point.}\label{fig_model}
\vspace{-10px}
\end{figure}

Our system is composed of a 1 DOF parallel gripper attached to a link that can rotate around a single axis. This system is an under-actuated two-link planar arm, in which the under-actuated joint corresponds to the pivoting point.  
We assume that we can control the desired acceleration on the first joint. This system is shown in Fig.~\ref{fig_model} above.

The dynamic model of the system is given by:
\begin{equation} \label{dynamic_model}
\begin{split}
(I+mr^2+mlr\cos(\phi_{tl}))\ddot{\phi}_{grp}+(I+mr^2)\ddot{\phi}_{tl}+... \\
...+mlr\sin(\phi_{tl})\dot{\phi}_{grp}^2+mgr\cos(\phi_{grp}+\phi_{tl}) = \tau_f,
\end{split}
\end{equation}
where the variables are as follows: 
$\phi_{grp}$ and $\phi_{tl}$ are the angles of the first and second link respectively; $\ddot{\phi}_{grp}$ and $\ddot{\phi}_{tl}$ are their angular acceleration and $\dot{\phi}_{grp}$ is the angular velocity of the first link; $l$ is the length of the first link; $I$ is the tool's moment of inertia with respect to its center of mass; $m$ is the tool's mass; $r$ is the distance of its center of mass from the pivoting point; $g$ is the gravity acceleration; $\tau_f$ is the torsional friction at the contact point between the gripper's fingers and the tool. 
In our case, the second link represents the tool and $\phi_{tl}$ is the variable we aim to control

On a real setup, the modeled two-link arm is the final part of a robotic manipulator. Therefore, the gravity component $g$ varies according to the current orientation of the plane that contains the actuated link and the tool. We assume that the manipulator's configuration is such that this plane has only pitch angle and no roll. As a result, the acceleration due to gravity influences only one direction of motion and $g$ varies between 0 and 9.8 according to the pitch angle, without the need of an additional term in Equation \ref{dynamic_model}.

\subsection{Friction Model}
Our proposed solution for pivoting exploits the friction at the contact point between the gripper and the tool to control the rotational motion. Such friction is controlled by enlarging or tightening the grasp.  

When the tool is not moving, that is $\dot{\phi}_{tl}=0$, the static friction $\tau_s$ is modeled according to the Coulomb friction model:
\begin{equation} \label{static_friction}
|\tau_s|\leq\gamma f_n,
\end{equation}
where $\gamma$ is the coefficient of static friction and $f_n$ is the normal force applied by the gripper's fingers on the tool.

When the tool moves with respect to the gripper, that is $\dot{\phi}_{tl}\neq0$, we model the friction torque $\tau_f$ as viscous friction and Coulomb friction \cite{olsson}:
\begin{equation}  \label{motion_friction}
\tau_f=-\mu_v\dot{\phi}_{tl}-\mu_cf_n\sgn(\dot{\phi}_{tl}),
\end{equation}
in which $\mu_v$ and $\mu_c$ are the viscous and Coulomb friction coefficients respectively and $\sgn(\cdot)$ is the signum function.

When the tool starts moving numerical singularities can occur due to switching between the two models in Equations \ref{static_friction} and \ref{motion_friction}. To alleviate this problem we follow the approach proposed in \cite{karnopp}. We define a neighborhood $|\dot{\phi}_{tl}|\leq\epsilon$, for a small $\epsilon>0$, where the friction torque is equal to the net torque acting on the tool. Therefore, when the tool has zero velocity, the normal force will counterbalance the net torque.

Since most of the robots are not equipped with tactile sensors to measure the normal force $f_n$ at the contact point, we follow the approach proposed in \cite{vina_1} and express this force as a function of the distance $d_{fing}$ between the two fingers, assuming a linear deformation model: 
\begin{equation}
f_n=k(d_0 -d_{fing}),
\end{equation}
where $k$ is a stiffness parameter and $d_0$ is the distance at which there is no fingertip deformation. In other words, $d_0$ is the distance at which the fingers initiate the contact with the tool.

%==========================================================
\section{Learning}
\label{sec_learning}

As discussed in Section~\ref{subsec_relatedwork_rlrobotics}, recent algorithmic advances in reinforcement learning suggest the possibility that model-free algorithms could be applied to robotics tasks. However, since these algorithms frequently are not designed to be sample-efficient enough to learn in real time on the hardware, our approach is to construct a training procedure to learn robust policies in a simple custom simulated environment, then deploy on the robot. 

Below we first describe the overall training approach, then present formalization of the pivoting task as a Markov Decision Process, then give a brief summary of the model-free reinforcement learning algorithm we used for policy search.

%------------------------------------------------------------------------------
\subsection{Learning Robust Policies using a Simulator}
\label{subsec_learning_sim}

As described in Section~\ref{sec_problem_description}, our approach is to enable learning from simulated environment, while being robust to the discrepancies between the simulation and actual execution on the robot. For this purpose, we first built a simple custom simulator using the equations described in Section~\ref{sec_system_model}. Then, to facilitate learning policies robust to uncertainty, we injected up to 10\% randomized delay for arm and finger actions in simulation. We also added 10\% noise to friction values estimated for the tool modeled by the simulator. We then trained a model-free deep reinforcement learning policy search algorithm on our simulated setting. Lastly, we executed the resulting policy on the robot (Baxter) for evaluation. This approach allowed us to keep the simulator simple and fast, while still enabling learning policies robust to the mismatch of the simulated and real environments.

%In order to make learning tractable for model-free algorithms that benefit from the ability to sample a large number of learning trajectories, 

%------------------------------------------------------------------------------
\subsection{Pivoting Task as a Markov Decision Process}
\label{subsec_learning_mdp}

We formulate the pivoting task as a Markov Decision Process (MDP) -- a tuple $\{S, \ A, \ P(s' | s, a), \ R(s,a,s'), \ \alpha \}$.

\noindent The state space $S$ is comprised of states $s_t$ observed at each time step $t$: $s_t\!=[ \phi_{tl} - \phi_{tgt} \ ,\ \dot{\phi}_{tl} \ ,\  \phi_{grp} \ ,\  \dot{\phi}_{grp} \ ,\  d_{fing} ]$, with notation as in Section~\ref{subsec_dynamic_model}.
%$\phi_{tl}, \dot{\phi}_{tl}$ is the angle and angular velocity of the tool with respect to the central axis of the gripper, $\phi_{tgt}$ is the target angle for the tool, $\phi_{grp}, \dot{\phi}_{grp}$ is the angle and angular velocity of the gripper (attached to the arm for the robot), and $d_{fing}$ is the distance between the fingers of the gripper. 
Using the signed distance of the tool angle as the first component of the state vector allows us to facilitate generalization in learning, since in our setting the motions of the robot arm are symmetric, e.g. the optimal policy for reaching target angle $0^{\circ}$ starting from tool at $-45^{\circ}$ would be symmetric to the case of starting from $45^{\circ}$. For the settings without the symmetry the state space representation could instead include $\phi_{tl}$ and $\phi_{tgt}$ separately.

The action space $A$ is comprised of actions $a_t$ at time $t$: $a_t=\{\ddot{\phi}_{grp}, d_{fing} \}$, where $\ddot{\phi}_{grp}$ is the rotational acceleration of the robot arm, and $d_{fing}$ is the distance between the fingers of the gripper. In cases where the hardware is limited in ability to achieve fingers' distances precisely, it is advantageous to learn to control the direction of the change in distance instead of precise target distance. We discuss this further in the experiments section. 

The transition probabilities $P(s' | s, a)$ are not used during learning explicitly, since we employ model-free approaches for learning. The dynamics of the state transitions is implemented by the simulator (as described in Section~\ref{subsec_dynamic_model}), but the learner does not have an explicit access to the state transition dynamics.

The reward function $R$ gives a reward $r_t \in [-1,1]$ at each time step $t$ such that higher rewards are given when the angle of the tool is closer to the target angle: $r(t) = \frac{- |\phi_{tl} - \phi_{tgt}|}{\phi_{RNG}}$, where $\phi_{RNG}$ is a normalizing constant denoting the range of angles the tool can attain, e.g. close to 2$\pi$ if the motion is not further restricted by the shape of the tool or the gripper. A bonus of 1 is given when the goal region is reached -- the tool is close to the target angle and the velocity of the tool with respect to the gripper is not changing, i.e. the tool is gripped firmly.

Finally, to obtain infinite horizon discounted MDP, since we aim to achieve the goal within 100 steps on average, we use the discount factor of $\alpha=0.99$ (from $\frac{1}{1-\alpha}\!=\!100$). Of course the horizon of 100 can be changed as needed depending on the desired duration of the pivoting task.

%------------------------------------------------------------------------------
\subsection{Policy Search using Reinforcement Learning}
\label{subsec_learning_trpo}

As discussed in Section~\ref{subsec_relatedwork_rlrobotics}, several continuous control model-free reinforcement learning algorithms would be suitable for learning optimal policies for our formulation of the pivoting task as MDP. In our experiments, we found that Trust Region Policy Optimization~\cite{schulman2015trust} was able to solve the problem without extensive parameter adjustment. Below we summarize the main ideas behind TRPO for a brief overview.

TRPO is a method for optimizing large non-linear control policies, as those represented by neural networks. The algorithm aims to ensure monotonic improvement during training by computing a safe region for exploration.

The main optimization performed by the algorithm is to iteratively solve a set of optimization problems:
\begin{align}
\label{eq_trpo}
\begin{split}
&\text{maximize}_{\theta} \ E_{s \sim \rho_{\theta_{old}}, \ a \sim q} \Bigg[ \frac{\pi_{\theta}(a|s)}{q(a|s)} Q_{\theta_{old}}(s,a) \Bigg] \\
&\text{subject to } E_{s \sim \rho_{\theta_{old}}} \big[ D_{KL}(\pi_{\theta_{old}}(\cdot | s) || \pi_{\theta}(\cdot | s)) \big] \leq \delta,
\end{split}
\end{align}
where $\theta_{old}$ denotes the initial (or previous) set of policy parameters, $\theta$ denotes the updated policy parameters, $\pi_{\theta}$ is the stochastic policy parameterized by $\theta$, $q$ is a sampling distribution for exploration, $Q_{\theta_{old}}$ is the Q function approximator estimated from previous samples, and the expectation $E$ is taken over samples obtained using the policy from previous iteration (see~\cite{schulman2015trust} for details). The constraint in the optimization aims to keep the new policy sufficiently close to the old to yield (in theory) monotonically improving policies by limiting KL divergence of $\pi_{\theta}(\cdot | s)$ from $\pi_{\theta_{old}}(\cdot | s)$.
Briefly, the overall structure of the algorithm is: 1) collect a set of state-action pairs along with Monte Carlo estimates of their Q-values; 2) by averaging over samples, construct the estimated objective and constraint from Equation~\ref{eq_trpo}; 3) approximately solve this constrained optimization problem to update policy parameter vector~$\theta$ (using conjugate gradient followed by a line search).

TRPO has been shown to be competitive with (and sometimes outperform) other recent continuous state and action RL algorithms~\cite{duan2016benchmarking}. However, to our knowledge it has not yet been widely applied to real-world robotics tasks. While the background for the algorithm is well-motivated theoretically, the approximations made for practicality, along with challenges in achieving reasonable training results with a small-to-medium number of samples, could impair the applicability of the algorithm to learning on the robot directly. Hence we explore the approach of using TRPO for policy search in a simulated environment.

%==========================================================
\section{Experiments}
\label{experiments}
\label{sec_experiments}
In this section we first discuss implementation details of training in the simulator. We then show initial evaluation results indicating potential for robustness to changes in friction. Then we describe experiments run on Baxter robot. We discuss the performance on the tool with parameters similar to the ones used during training, as well as results for manipulating the tool with different shape, mass, inertia parameters and unknown friction properties. The results indicate that our training procedure is able to produce robust policies capable of solving the pivoting task on both known and unknown tool.

%------------------------------------------------------------------------------
\subsection{Learning Robust Policies in Simulation}
\label{subsec_experiments_learning}

To facilitate experimenting with various RL algorithms, we implemented a custom environment in OpenAI Gym~\cite{OpenAIGym} for the pivoting task MDP (as defined in Section~\ref{subsec_learning_mdp}). To experiment with TRPO~\cite{schulman2015trust} and DDPG~\cite{lillicrap2015continuous} algorithms, we used rllab implementation~\cite{duan2016benchmarking} as a starting point, then adjusted the behavior and parameters as needed for various experiments for this project. Both TRPO and DDPG papers documented exploration and network training parameters used to obtain results for simulated control tasks. Starting with these reported values, we experimented with several options like rate of exploration, batch size for neural network training and network size. We were not able to obtain robust learning with DDPG (training progress would frequently stall), so for the rest of the experiments we decided to only use TRPO, which in our setting exibited a gradual but satisfactory learning progress.

\begin{figure}[H]
\centering
\caption{\small{Evaluation of the training progress of TRPO algorithm. 50~episodes were used for each training iteration. The goal region was~$\pm3^{\circ}$ of the target angle. Left plot shows average reward when evaluating the policy obtained at the current training iteration on angles randomly selected from $[-\pi/2.5, \pi/2.5] \ (\approx~\!~[-72^{\circ}, 72^{\circ}])$. Right plot shows average number of steps before reaching the goal (with a cap of 250 steps per episode).}}
\label{fig_train_progress}
\includegraphics[width=0.49\textwidth]{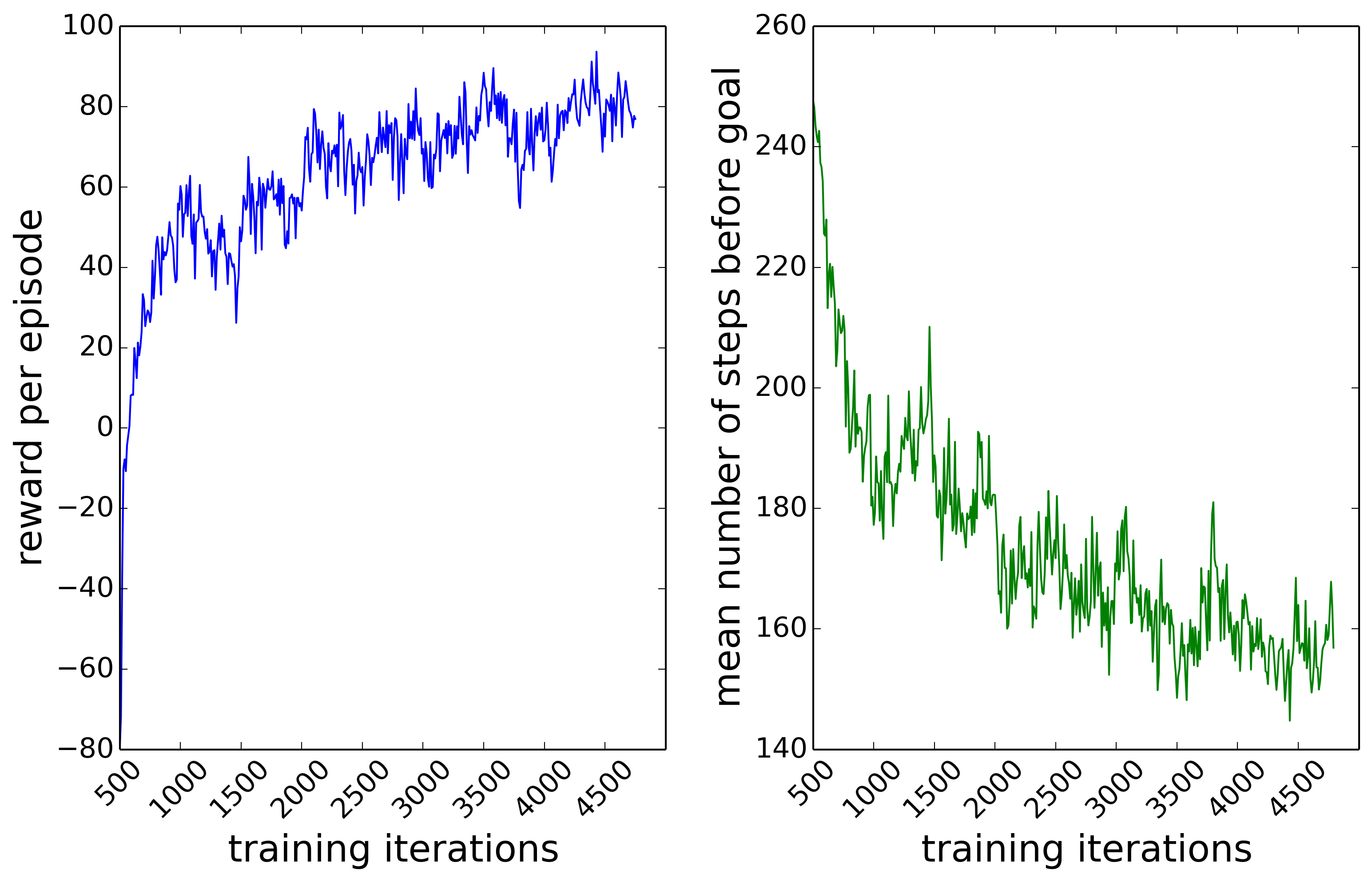}
\end{figure}

We trained TRPO with a fully connected network with 2 hidden layers (with 32, 16 nodes) for policy and Q function approximators. We also experimented with larger networks of up to 3 hidden layers, though found that the smaller network was enough to solve the task. When generating training episodes initial and target angles were chosen at random from $[-\pi/2,\pi/2]$; each training episode had 100 steps. The motion was constrained to be linear in a horizontal plane. However, since the learning was not at all informed of the physics of the task, any other plane could be chosen and implemented by the simulator if needed. Fig.~\ref{fig_train_progress} below visualizes evaluation of the policies as the number of training iterations increases.

To simulate the dynamics of the gripper arm and the tool we used the modeling approach described in Section~\ref{sec_system_model}. We used the parameters of tool~1 as specified in Table~\ref{tool_parameters} with friction coefficients from Equation~\ref{motion_friction} set to $\mu_v~=~0.066, k \mu_c~=~9.906$. 

\begin{figure}[H]
\centering
\caption{\small{Evaluating performance on friction coefficients outside of the noise range used during training. Coulomb friction was set to $k \mu_c=9.906$ during training, with noise of up to $\pm 0.1 k \mu_c$. The plots below show the average reward and \% of evaluation episodes with goal reached when evaluated on settings with friction coefficient significantly larger or smaller than the range used during training. Goal region, initial and target angles same as in Fig.~\ref{fig_train_progress}.}}
\label{fig_sim_fric_robust}
\includegraphics[width=0.49\textwidth]{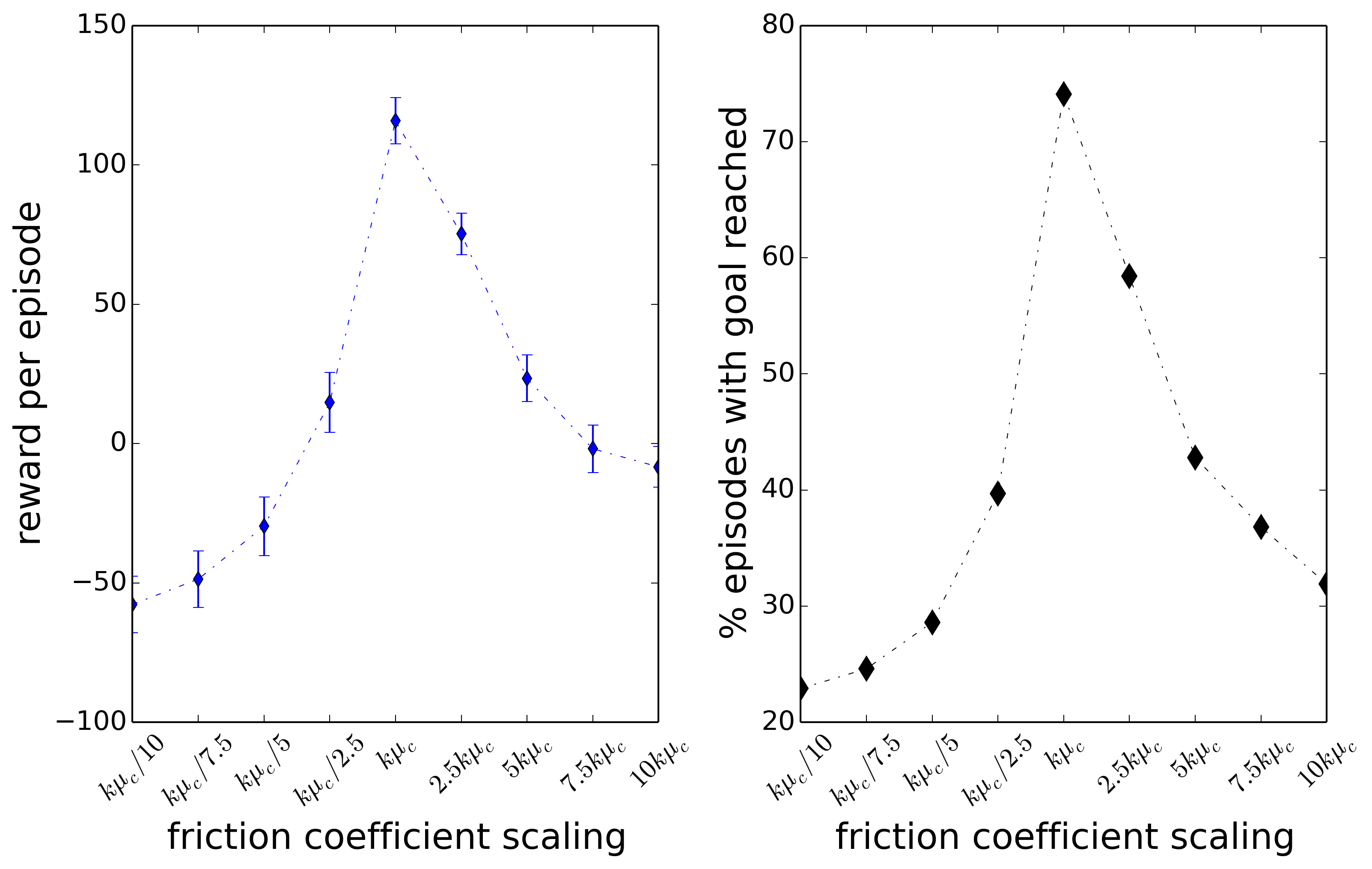}
\end{figure}

As discussed in \mbox{Section~\ref{subsec_learning_sim}}, the noise injected when training in simulation aimed to help learning policies robust to mismatch between the simulator and the robot. Fig.~\ref{fig_sim_fric_robust} below visualizes using the policy obtained after 20K training iterations on simulated settings with a mismatch larger than noise used during training. While we allowed only up to 10\% noise in the value $k \mu_c$ which modeled Coulomb friction, we observed that the policy was still able to retain 40\% success rate of reaching the target angle when tested on settings with 250\% to 500\% change in $k \mu_c$. This level of robustness suggests we could to avoid estimating friction coefficients precisely. This is crucial for the pivoting task, since, unlike estimating the dimensions of the tools, estimating the friction properties precisely would be an intractable challenge for most widely available robotic systems.

%------------------------------------------------------------------------------
\subsection{Experiments on Hardware with Baxter}
\label{subsec_experiments_baxter}

We implemented the proposed approach on a Baxter robot, using one of its 7 DOF arms. We selected slightly deformable fingertips for the gripper to be able to control the force applied on the tool by changing the distance between the fingers. As for many commercial robots, Baxter's gripper cannot be controlled at the same high frequency as the joints. In particular, we measured that the gripper's fingers were able to reach a desired distance only after allowing a delay of $80\mathrm{ms}$. This delay entails longer time needed for executing each control action on the hardware, hence longer times needed by the tool to reach the goal due to these limitations of the robot.

To estimate the angle $\phi_{tl}$ we used a color based segmentation to track a colored marker on the tool. The images were collected by a Kinect 2 RGB-D camera running at $30~\mathrm{fps}$.

During the experiments, we kept the gripper and the actuated joint in a horizontal position, such that the gravitational acceleration would be $g=0$ and the motion of the tool was only determined by the inertial forces generated by the commanded actions. The distance between the actuated joint and the pivoting point was $l=0.35~\mathrm{m}$.

We performed experiments with two different objects of different materials, hence different friction properties. The parameters of the tools are shown in Table \ref{tool_parameters}. 

\begin{table}[H]
\caption{Parameters of the tools.}\label{tool_parameters}
\centering
\begin{tabular}{| @{}l@{} | l | c | c | c | c |} 
\hline
\multirow{ 3}{*}{
\centering
\includegraphics[width=0.123\textwidth]{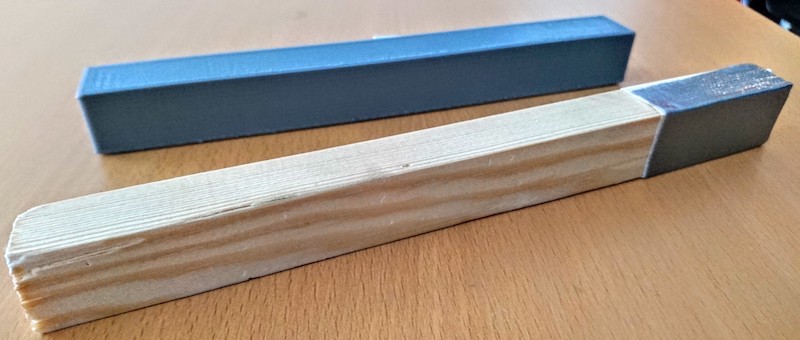}} &  & $I \mathrm{[kg\cdot m^2]}$ & $m \mathrm{[kg]}$ & $r \mathrm{[m]}$ & $d_0 \mathrm{[m]}$ \\\cline{2-6}
 & tool 1 &  0.00006943 & 0.026 & 0.089 & 0.0188 \\\cline{2-6}
 & tool 2 & 0.0001111 & 0.033 & 0.1 & 0.0162  \\\hline
\end{tabular}
\end{table}

The robot is able to estimate the difference in the finger's distance when closing them to grasp the tool and therefore it can estimate the difference in the $d_0$ value also for an unknown object.
The friction coefficients used for modeling the tool in the simulator have been estimated only for the first tool. Hence the policy was not explicitly trained using the parameters matching those of the second tool. 

Table~\ref{tool_trials} summarizes the results of our experiments on the robot. To streamline the experiments, we cycled through the following target angles: starting from $0^{\circ}$, reach $45^{\circ}$, then  $0^{\circ}$, then $-60^{\circ}$, then $30^{\circ}$, then $5^{\circ}$ and finally $0^{\circ}$. This allowed us to test the policy on both wide and narrow ranges of motion. With this we obtained a $\approx\!\!93\%$ success rate with tool~1 and $\approx\!\!83\%$ with tool~2. As expected, the policy performs better with the tool whose parameters were used (in a noisy manner) during training in simulation. However, it is important to note that the performance using the tool not seen during training was robust as well. The experiments on the robot were performed without re-adjusting the tool. As a consequence, we observed that eventual sliding of the tool would cause drops after several rounds of experiments. However, these were still very infrequent.

\begin{table}[H]
\centering
\caption{Results of hardware experiments on Baxter.}\label{tool_trials}
\begin{tabular}{| l | c | c |} \hline
 & tool 1 & tool 2 \\ 
 \hline
Goal reached & 28 & 25 \\
Goal not reached & 1 & 3  \\
Tool dropped & 1  &  2 \\
\hline
Total & 30 & 30 \\
\hline
\end{tabular}
\end{table}

Fig.~\ref{fig_train_progress_one} below illustrates two example episodes run on Baxter with the first and second tool. We observe that the target is reached faster when tool~1 is used (after $\approx 5s$), and a bit slower when tool~2 is used (after $\approx 10s$).

\begin{figure}[H]
\centering
\caption{\small{Two example trajectories from experiments on Baxter robot: reaching the target of $-60^{\circ}$ ($\pm3^{\circ}$ for goal region) from $0^{\circ}$. \\ Left: using tool~1 whose parameters are used in the simulator. \\ Right: using tool~2 whose parameters are not used for training.}}
\label{fig_train_progress_one}
\includegraphics[width=0.49\textwidth]{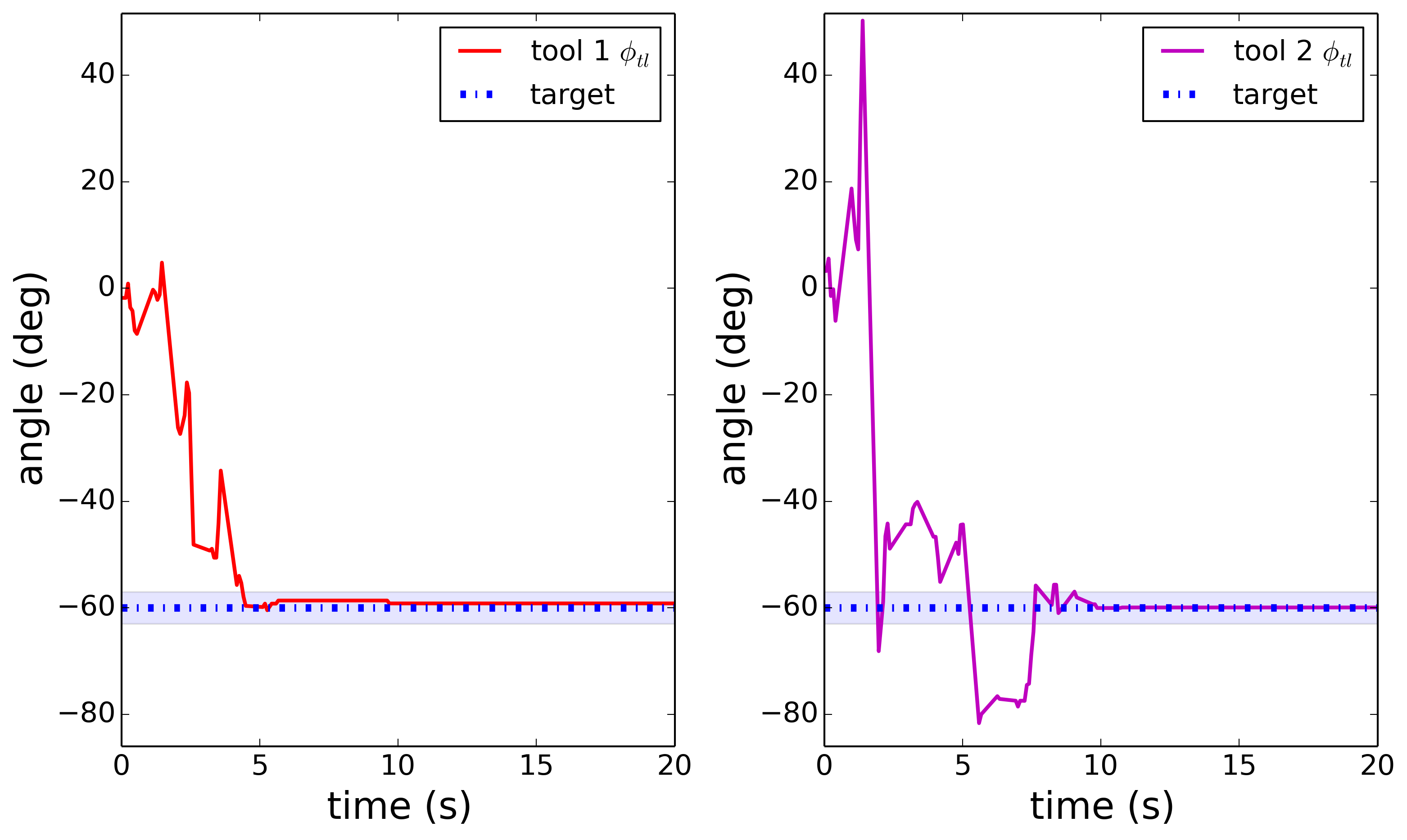}
\end{figure}

Fig.~\ref{fig_train_progress_all} below illustrates the performance averaged over all the trials. The deviation from the goal region after reaching the goal is likely due to inaccuracies in tracking. After reporting that the tool is in the goal region, the tracker might later report a corrected estimate, indicating further tool adjustment is needed. We observe that in such cases the policy still succeeds in further pivoting the tool to the target angle.

\begin{figure}[H]
\vspace{-5px}
\centering
\caption{\small{Mean distance to target vs time for experiments on Baxter averaged over all trials (excluding drops, since tool angle is not tracked after a drop). Left: using tool~1. Right: using tool~2.}}
\label{fig_train_progress_all}
\includegraphics[width=0.48\textwidth]{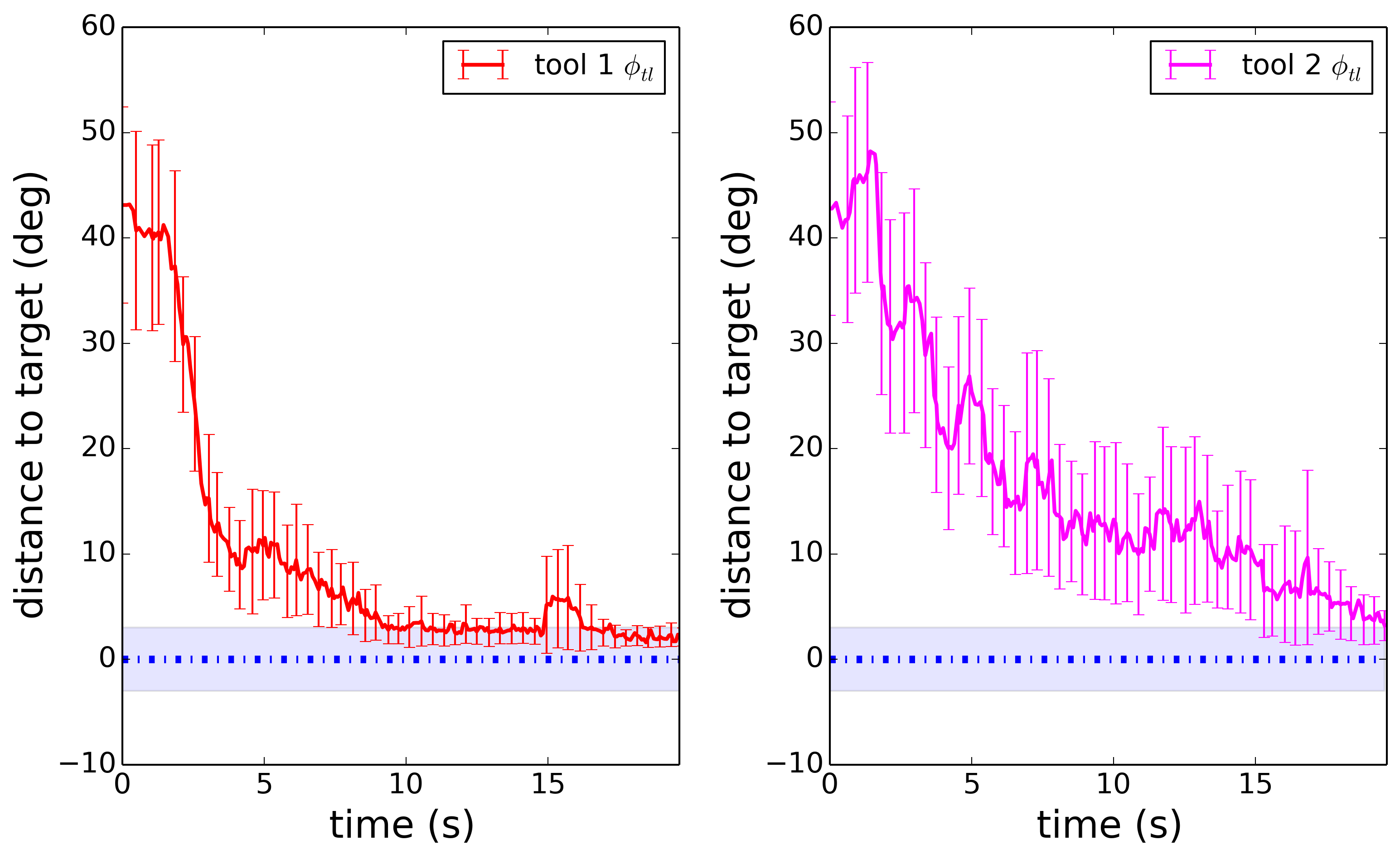}
\vspace{-20px}
\end{figure}

%==========================================================
\section{Conclusions and Future Work} 
\label{sec_conclusions}
In this work we proposed solving the pivoting task via building a simple custom simulator of the task dynamics and using reinforcement learning to learn control policies. We presented the learning procedure that is able to manage the mismatch between the simulated and real settings, hence does not require precise estimates of all the tool parameters. We then demonstrated the ability of the learned policy to solve the pivoting task on the Baxter robot.

\vspace{5px}

To extend our work to a more general set of dextrous manipulation problems, in the future we can combine pivoting with sliding. This would enable more possibilities for repositioning. In fact, with pivoting we can change the orientation of the object around a single axis, while with sliding we can translate the object, hence changing the contact point. The next step would be to incorporate existing approaches to modeling sliding into the simulator.

We can also explore augmenting the learning part of our approach. In situations where further training on the hardware is possible, we could develop a second stage to fine-tune the policies learned in simulation. This could be accomplished, for example, by quickly learning an adjustment to the discrepancy between the simulated and real environment without the need to re-learn control policies from scratch. Another direction is to experiment with recurrent networks to model aspects of the environment not observable directly.
%instead of limiting the formulation of the learning problem to an MDP.

Further into the future we can consider the very challenging problem of manipulating non-rigid objects. Building high fidelity simulators for deformable objects and objects with variable center of mass has been mostly intractable in the past. However, approximate simulators can be constructed. We have observed that for rigid objects only approximate simulation of the dynamics is sufficient to learn effective control policies. So we can explore whether our approach with learning from approximate simulation can be adapted to the case of manipulating deformable objects.

%%%%%%%%%%%%%%%%%%%%%%%%%%%%%%%%%%%%%%%%%%%%%%%%%%%%%%%%%%%%%%%%%%%%%%%%%%%%%%%%

\section*{ACKNOWLEDGMENT}
This work was supported by the European Union framework program H2020-645403 RobDREAM. \\

\vspace{-10px}

%%%%%%%%%%%%%%%%%%%%%%%%%%%%%%%%%%%%%%%%%%%%%%%%%%%%%%%%%%%%%%%%%%%%%%%%%%%%%%%%

%%%%%%%%%%%%%%%%%%%%%%%%%%%%%%%%%%%%%%%%%%%%%%%%%%%%%%%%%%%%%%%%%%%%%%%%%%%%%%%%
\bibliographystyle{IEEEtran}
\bibliography{Pivoting,DexterousManip,FrictionControl,RLinRobotics,General}

\end{document}